\documentclass[letterpaper, 10 pt, conference]{ieeeconf}  
\usepackage[table]{xcolor}
\usepackage{graphicx}
\usepackage{booktabs}
\usepackage[T1]{fontenc}
\usepackage[utf8]{inputenc}
\usepackage[english]{babel}
\usepackage{multirow}
\usepackage{multicol}
\usepackage{caption}
\usepackage{subcaption}
\usepackage{amsmath}
\usepackage{amssymb}
\usepackage{capt-of}
\usepackage[export]{adjustbox}
\usepackage{pifont}

\usepackage{makecell}

\usepackage[colorlinks,pagebackref=false,citecolor=green,bookmarks=false,hypertexnames=true]{hyperref}

\IEEEoverridecommandlockouts                              

\title{\LARGE \bf
Exploring 2D Data Augmentation for 3D Monocular Object Detection
}

\author{Sugirtha T$^{1}$, Sridevi M$^{1}$, Khailash Santhakumar$^{2}$, B Ravi Kiran$^{3}$,  Thomas Gauthier$^3$ and Senthil Yogamani$^{4}$ \\
$^{1}$NIT Tiruchirappalli, India \quad
$^{2}$SASTRA University, India \quad
$^{3}$Navya, France \quad
$^{4}$Valeo, Ireland
}

\begin{document}


\twocolumn[{
	\renewcommand\twocolumn[1][]{#1}
	\maketitle
	\begin{center}
		\vspace{-0.6cm}
		\includegraphics[width=\textwidth]{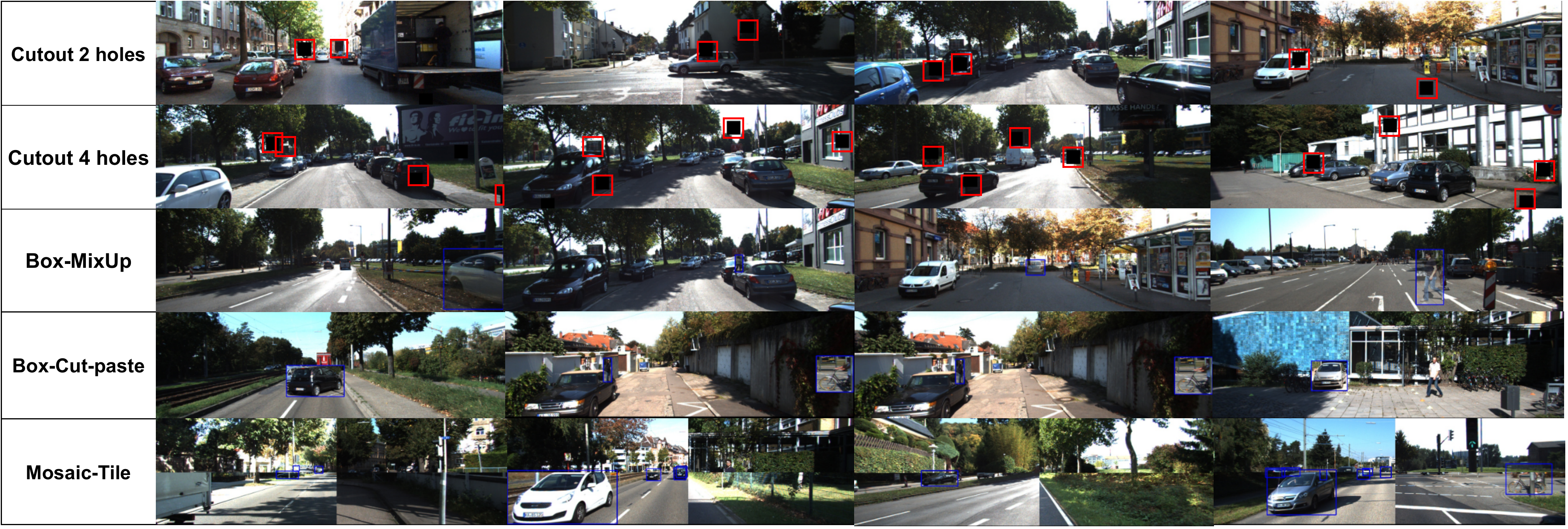}
		\captionof{figure}
		{
			Illustration of of different 2D data augmentations for Monocular 3D Object Detection. Red boxes highlight cutout regions and blue boxes denote object annotations which are used to mix or paste.
		}
		\label{fig:teaser}
	\end{center}
}]

\begin{abstract}

Data augmentation is a key component of CNN based image recognition tasks like object detection. However, it is relatively less explored for 3D object detection. Many standard 2D object detection data augmentation techniques do not extend to 3D box. Extension of these data augmentations for 3D object detection requires adaptation of the 3D geometry of the input scene and synthesis of new viewpoints. This requires accurate depth information of the scene which may not be always available. In this paper, we evaluate existing 2D data augmentations and propose two novel augmentations for monocular 3D detection without a requirement for novel view synthesis. We evaluate these augmentations on the RTM3D detection model firstly due to the shorter training times . We obtain a consistent improvement by 4\% in the 3D AP (@IoU=0.7) for cars, ~1.8\%  scores 3D AP (@IoU=0.25) for pedestrians \& cyclists, over the baseline on KITTI car detection dataset. We also demonstrate a rigorous evaluation of the mAP scores by re-weighting them to take into account the class imbalance in the KITTI validation dataset.


\end{abstract}

\section{Introduction}
3D object detection is crucial perception task in modern autonomous driving applications, used upstream for scene understanding, object tracking and trajectory prediction and decision making. Initially, autonomous cars are equipped with LiDAR sensors and most 3D detectors rely on LiDAR data to perform 3D object detection. LiDAR provides precise distance measure which makes it feasible to detect accurate 3D bounding boxes. But, they are expensive to be deployed in autonomous cars. Recent autonomous cars use single monocular camera and hence monocular 3D object detection (3D OD)became a research focus in computer vision community.
3D OD methods are usually part of the 4 following categories: (1) 2D proposal generation \cite{Law_2018_ECCV}, (2) Geometric constraints \cite{Zhou_2019_CVPR}, (3) Keypoints detection \cite{Zhou2019ObjectsAP} or (4) Direct 3D proposals generation \cite{Li2020Monocular3D}. 

Convolutional Neural Networks (CNNs) learn complex hierarchical features from images. However, the models are susceptible to overfitting and require proper regularization techniques to be applied. Usual regularization techniques include data augmentation, ie. image transformations such as scaling, translation, random flipping etc. In computer vision, data augmentation is pervasive due to its simplicity to implement and efficiency. In recent years, few regularization methods have been proposed to reduce overfitting of the model while training. This also improves the robustness of CNNs to learn complex features.

RTM3D\cite{rtm3deccv2020} is a monocular 3D object detector that only applies flipping to augment the training samples. In this paper, we propose variants of 2D data augmentations such as Cutout, CutPaste and Mixup and try all possible combinations to find out the best suited combination in order to improve the detection accuracy over the baseline RTM3D detector model.
We define a family of 2D data augmentations and their transformations on monocular images for 3D object detection. Also, we perform an ablation study on which our proposed augmentation methods improve the performance over the RTM3D baseline. 3D monocular detection requires an expensive annotation process. 3D detection methods have mostly been trained on large-scale datasets, while data augmentation methods do exist but have not been explored further as in case of YOLOv3\cite{Redmon2018YOLOv3AI} or modern object detectors. Our new data augmentation methods improve performance greatly on 3D object detection for occlusion, pixel corruption and better localization.

This paper evaluates an initial set of 2D data augmentation techniques for monocular 3D detection, without changing the 3D geometry of bounding boxes in the scene and avoiding the requirement to synthesize new viewpoints of the input camera view.
In summary, the contributions of our paper are as follows:
\begin{itemize}
    \item We evaluate the performance of photometric as well as region based 2D data augmentations for the monocular 3D object detection task.
    \item We propose three new 2d data augmentation strategies that significantly improve the detection accuracy of the baseline RTM3D detector namely Box-MixUp, Box-Cut-Paste, and Mosaic-Tile.
\end{itemize}
Extensive analysis on KITTI dataset\cite{kittidataset2012} demonstrates that our proposed augmentations improve performance under various conditions: occlusions, contrasted/shadowed pixels, changing the diversity of viewpoints of objects seen in the dataset.

\section{Related Work}

2D OD on image plane is inadequate for reliable autonomous driving scenario because it does not provide an accurate estimation of 3D objects sizes and space localization. In other words, 2D OD methods have limited performance in following scenarios namely  Occlusion,  Object pose estimation and 3D position information. A 3D bounding box provides precise information about size of the object and its position in 3D space.\\

\noindent
\textbf{3D Object Detection :}
 Recently, many researchers proposed various techniques for 3D Monocular Object Detection. RTM3D\cite{rtm3deccv2020} and its extension KM3D\cite{Li2020Monocular3D} use CenterNet\cite{Zhou2019ObjectsAP} to regress a set of 9 projected keypoints corresponding to a 3D cuboid in image space (8 vertices of the cuboid and its center). They also perform direct regression for the object's distance, size and orientation. These values are then used for offline initialization of an optimizer to estimate 3D bounding boxes under geometric constraints.
 CenterNet provides basic data augmentation such as affine 
 transformations (shifting, scaling) and random horizontal flipping. Over
 these, KM3D adds coordinate independent augmentation via random color
 jittering. 

SMOKE\cite{liu2020smoke} regresses 3D bounding box directly from image plane which eliminates 2D bounding box regression. It represents an object by a single keypoint and these keypoints are projected as 3D center of each object. Mono3D\cite{Chen_2016_CVPR} is a region proposal based method that uses semantics, object contours and location priors to generate 3D anchors. It generates proposal by performing exhaustive search on 3D space and use Non-maximal suppression for filtering. SMOKE augments the training samples with horizontal flipping, scaling and shifting. GS3D\cite{Li_2019_CVPR} predicts the guidance of cuboid and performs feature extraction by projecting region of guidance. GS3D performs monocular 3D detection without augmenting the training data.
We observe that few detectors applied basic augmentations and few others applied none. In this paper, we propose data augmentations that can be deployed on the 3D detectors to improve their performances. \\

\noindent
\textbf{Advanced 2D Data Augmentations :}
Motivated by dropout regularization technique, few 2D data augmentations were proposed in recent years for object classification task. \textbf{CutOut}\cite{Devries2017ImprovedRO} removes a square region from the input image and fills it with any of the following : (i) black pixels (ii) grey pixels (iii) Gaussian noise. In other words, it performs dropout but in the input space. This drives the model to spotlight on the whole image instead of focusing on a few key features. Still, there is a chance to loose informative pixels during training.
\textbf{Mixup}\cite{Zhang2018mixupBE} performs linear interpolation of two randomly drawn images. Yet, the resultant images sometimes look unusual and it may confuse the model during localization task.

\textbf{CutMix}\cite{yun2019cutmix} overcomes the limitations in CutOut and Mixup by replacing the square region with a patch of identical dimensions from another image. This results in no loss of informative pixels while training, besides preserving the advantages of regional dropout in input space. Also, the images formed are legitimate compared to Mixup.
\textbf{Smoothmix}\cite{lee2020smoothmix} is proposed to minimize the “strong-edge” problem caused due to regional dropout. The model takes two images from training set and generates augmented images by performing element-wise addition of two images $(x_i,x_j)$ after an element-wise multiplication with a smoothly transitioning mask G.

All 2D data augmentations discussed above when applied to 3D space require synthesizing new view points of the input view and orientation of the bounding boxes will get distorted. In this paper, we extend existing 2D data augmentations without modifying the 3D geometry of bounding box and avoid synthesizing new view points for the input camera view.\\

\noindent YOLOv4\cite{Bochkovskiy2020YOLOv4OS} is a 2D object detector which which has improved object detection accuracy mainly thanks to adequate data augmentation such as Cutout, Mixup and CutMix. Additionally, they introduced the mosaic data augmentation to detect small objects. Mosaic data augmentation in YOLOv4 merges 4 training samples directly and applies random cropping to produce augmented image.
The recent developments in 2D data augmentations suggest their potential utility for the monocular 3D detection task, while assuming some geomtrical constraints.

\begin{table*}[t]
\caption{\label{tab:data_aug2d} List of data augmentations studied for 3D-MOD. $^*$ indicates proposed 2D transforms.}
\centering
 \begin{tabular}{l p{10cm}} 
 \toprule
Data augmentation Transforms & Description and Utilization \\ \midrule
 Cutout (holes, W)       &  Cutout removes a randomly chosen patch from the image and those pixels are zeroed. Cutout parameters include the number of square holes and their side.\\
 Photometric pixel-wise DA & Standard pixel level data augmenations which apply motion blur, RGB shifts and Random contrast from the albumentations library \cite{albumentations2020mdpi}. \\

 Box - MixUp$^{\ast}$   &   Generates an augmented (image, bounding box set) pairs by averaging the pixel values  between the  two images (reference \& random samples) only within masks defined by the bounding box. \\
 Box - Cut - Paste$^{\ast}$ & A variation of Box-Mixup is local-cut-paste where regions of the 2D bounding box from the second image is pasted on to the reference image. \\
 Mosaic-Tile$^\ast$ & Augmented image is created by tiling/concatenating the 4 corresponding quadrants of four different training samples into one image.\\
\bottomrule
 \end{tabular}
 \vspace{5pt}
 \end{table*}
 
\noindent
\textbf{3D augmentations with view synthesis:}
Both traditional and advanced data augmentations discussed above can be applied only in 2D image space.
To handle data augmentations for objects in 3D, performing in varying viewpoints a view synthesis problem is required to be solved.
RoI-10D\cite{manhardt2019roi} lifts 2D RoIs to 3D for 6 DoF pose regression. Authors show that 3D synthetic data augmentation was achieved by in-painting 3D meshes directly onto the 2D scenes.
MoCa - multi modality cut and paste\cite{Zhang2020MultiModalityCA} was proposed to overcome the limitations of cut-paste\cite{Dwibedi2017CutPA} that is applicable for single-modality 3D detectors. MoCa is readily applicable for multi-modality 3D object detection while preserving consistency when fusing point cloud and image pixels. 
Dense LiDAR pre-mapping can be leveraged to aid cutting out dynamic objects \cite{ravi2018real}.


\section{Proposed Method}
Generally there are multiple families of 2D data augmentations that are employed in literature discussed below.

\textbf{Pixel level data augmentation :} 
Photometric augmentation includes changes to the pixel values such as contrast, blurring, brightness or color changes.

\textbf{Regional data augmentation :}  We refer to data augmentations that mask, transform or change multiple locations in the image domain at the same time as regional data augmentations.
Coarsedrop \cite{albumentations2020mdpi} is an example of region transform. Occlusion occurs because of
regional augmentation and our proposed Cutout helps the detector to 
detect occluded objects more accurately. Hence it overcomes the effects 
due to region based augmentations and thus increasing the robustness of 
the detector. Motion blur is simulated by the blurring transformation which is regional augmentation.

\textbf{Geometric data augmentation :} transfers image pixels to new positions and hence changes the image geometry. This can include reflection, rotation, cropping, translation, scaling, flipping, etc.  


\subsection{Data augmentation (DA) methods}
The input training image is represented by tensor $ x \in {R}^{W \times H \times 3}$ while the training bounding boxes coordinates by $y$ along with its class label. The data augmented samples are represented by $(\widetilde{x}, \widetilde{y})$. They are generated from reference images designated by $(x_A, y_A)$ and $(x_B, y_B)$ (image, bounding box set). 
The data augmentations we proposed in this paper are under one of the two categories:

\begin{itemize}
    \item Data augmentations that transform only the input training image $x_A$ without modifying the ground truth boxes $y_B$, Ex. Cutout and Photometric transformations.
    \item Augmentations that modify both image and ground truth boxes. Ex. Box-Mixup
\end{itemize}

Finally, let $M \in \{0, 1\}^{W \times H}$ be a binary mask which is 1 for pixels included by the set of bounding boxes in ground truth $y$, and 0 otherwise. The list of data augmentations evaluated in this study are described in Table \ref{tab:data_aug2d} and illustrated in Fig. \ref{fig:teaser}.\\

\begin{figure*}[t]
\begin{center}
\includegraphics[width=0.95\textwidth]{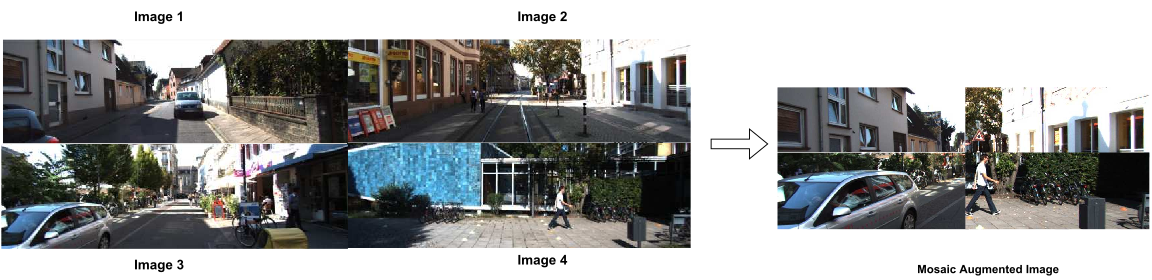}
\caption{Proposed mosaic data augmentation that preserves relative positions of 3D bounding box locations within each time as well as the final resolution of images.}
\label{fig:proposed-mosaic}
\end{center}
\end{figure*}

\textbf{Box-Mixup :}
Motivated by the work on Mixup data augmentation \cite{Zhang2018mixupBE} which was applied to image level classification problems, we aim here to propose an equivalent for 2D object detection. Box-MixUp is thus proposed to augment an image with object patches from other images, thus providing the same advantages of MixUp but localized over multiple regions in the image.
Box-MixUp generates an augmented sample pair by averaging pixels of two input images under a mask containing all bounding boxes from the second image, while concatenating both ground truth bounding box sets.
\begin{align}
\begin{split}
\widetilde{x} & = (0.5 x_A + 0.5 x_B) \cdot M_B + x_A \cdot (1-M_B)\\
\widetilde{y} & = y_A \cup y_B 
\end{split}
 \end{align}

\textbf{Box-Cut-Paste :}
Cut-Paste data augmentation \cite{Dwibedi2017CutPA} uses pre-existing masks and background scenes to artificially create an augmented dataset to train an object detection model. In Box Cut-Paste, we propose not to use a separate mask. The transformation directly pastes pixels under a bounding box from a reference image onto another image. This further ensures the object context is preserved and avoids unrealistic scenarios possible in Cut-Paste, such as cars in the sky. This is because the cut paste is only done over bounding box masks already existing in source images from the dataset. There might be cases where objects could get pasted on walls or other unusual backgrounds, though the perspective of the object remains natural.  
\begin{align}
\begin{split}
\widetilde{x} & = x_A \cdot (1-M_B) + x_B \cdot M_B \\
\widetilde{y} & = y_A \cup y_B  
\end{split}
 \end{align}

\textbf{Mosaic-Tile :} 
YOLOV4 proposed by authors \cite{Bochkovskiy2020YOLOv4OS} introduces the Mosaic augmentation, which takes 4 training images in different contexts and tiles them into one. It then performs random cropping to produce an augmented image of size equal to the original training image size. The batch normalization evaluates activation values across 4 different images reducing the requirement for a large mini-batch size. Mosaic increases the scale diversity of objects by tiling images at different scales.

Our proposed Mosaic-Tile augmentation takes distinct parts from four different training images and combines them into one as shown in Fig. \ref{fig:proposed-mosaic}. To ensure that we do not select bounding boxes from an input sample, which is outside its corresponding tile's extent in the augmented image, we ensure that at least 40\% of boxes area is within the corresponding tile. Ground truth boxes under this threshold and outside their tiles are eliminated from training to avoid over-fitting (insufficient image pixels available for these boxes). In future work this threshold hyper-parameter shall be studied to obtain the best detection results.
The advantage of our proposed Mosaic-Tile are quite similar to the original Mosaic augmentation. In our current evaluation we do not randomize the image coordinates at which we split the image and keep this to be the center of the image across all training. Here again, this a way to somehow preserve the scene geometry while combining different contexts.

\textbf{IoU check : } The Box-Mixup \& Box Cut-Paste augmentations were evaluated under two different conditions. 
\begin{itemize}
    \item In the first case we enforce that no two boxes between the two input samples overlap by a large margin. This was achieved by evaluating the 2D-IoU between pairs of boxes coming from two different samples and ensuring the intersection was smaller than 0.4. That is IoU$(B_i, B_j), \forall B_i \in y_A, B_j \in y_B$. If the IoU is too high, the incoming bounding box is rejected from the augmentation and the mixup/paste operation is not performed in the image domain.
    \item In the second case, we do not filter boxes using any IoU threshold and permit all possibilities.
\end{itemize}

In Table \ref{tab:mAP} the presence of IoU condition is designated by \textit{B+Box-Mixup (w iou)} and its absence \textit{B+Box-Mixup (w/o iou)}.

\section{Results}

\subsection{Baseline Model}
RTM3D\cite{rtm3deccv2020} is a real-time network that is based on the CenterNet architecture, which enables both fast training cycles and small inference times. Authors have already provided a  set of baseline data augmentations which include flip, affine transformations, stereo dataset augmentation using the left/right images in the KITTI dataset.
RTM3D architecture enables a faster evaluation of data augmentation strategies. It is important to note that the data augmentation methods proposed in this study are quite generic and can be employed to any other Monocular 3D object detection model. RTM3D and few other models had begun evaluating a simple set of 2D data augmentations, while there was no systematic study yet done, and was the motivation for this study.

\subsection{Dataset and Implementation Details}

\begin{table*}[htb!]
\tiny
\caption{\label{tab:results} {\textbf{AP$_{3D}$}} scores for data augmentations with RTM3D evaluated over : Car, Pedestrain and Cyclist. For each data augmentation method we show the difference in AP scores w.r.t the baseline. Green refers to positive gains, while red refers to negative drops in performance. The highest improvements have been highlighted for each difficultly level and class.}
\begin{tabular}{|l|c|c|c|c|c|c|c|c|c|c|c|c|c|c|c|c|c|c|}
\hline
\textbf{Classes}                  & \multicolumn{6}{c|}{\textbf{Car(IoU=0.7)}}                                                                                                                                                                                                    & \multicolumn{6}{c|}{\textbf{Pedestrian(IoU=0.25)}}                                                                                                                                                                                          & \multicolumn{6}{c|}{\textbf{Cyclist(IoU=0.25)}}                                                                                                                                                                                               \\ \hline
                                  & \multicolumn{3}{c|}{\textbf{AP$_{3D}$}}                                                                                    & \multicolumn{3}{c|}{\textbf{AP$_{BEV}$}}                                                                                   & \multicolumn{3}{c|}{\textbf{AP$_{3D}$}}                                                                                   & \multicolumn{3}{c|}{\textbf{AP$_{BEV}$}}                                                                                  & \multicolumn{2}{c|}{\textbf{AP$_{3D}$}}                                            & \textbf{}                             & \multicolumn{3}{c|}{\textbf{AP$_{BEV}$}}                                                                                   \\ \hline
\textbf{Augmentations}             & E                                     & M                                     & H                                     & E                                     & M                                     & H                                     & E                                    & M                                     & H                                     & E                                     & M                                     & H                                    & E                                     & M                                     & H                                     & E                                     & M                                     & H                                     \\ \hline
\textbf{Baseline (B)}             & \cellcolor[HTML]{CCCCCC}12.28            & \cellcolor[HTML]{CCCCCC}9.7              & \cellcolor[HTML]{CCCCCC}8.12             & \cellcolor[HTML]{CCCCCC}18.74            & \cellcolor[HTML]{CCCCCC}14.13            & \cellcolor[HTML]{CCCCCC}13.33            & \cellcolor[HTML]{CCCCCC}17.64           & \cellcolor[HTML]{CCCCCC}17               & \cellcolor[HTML]{CCCCCC}16.26            & \cellcolor[HTML]{CCCCCC}20.81            & \cellcolor[HTML]{CCCCCC}17.42            & \cellcolor[HTML]{CCCCCC}16.6            & \cellcolor[HTML]{CCCCCC}23.02            & \cellcolor[HTML]{CCCCCC}16.27            & \cellcolor[HTML]{CCCCCC}14.34            & \cellcolor[HTML]{CCCCCC}23.64            & \cellcolor[HTML]{CCCCCC}16.57            & \cellcolor[HTML]{CCCCCC}14.39            \\ \hline
\textbf{B+cutout 2 holes}         & \cellcolor[HTML]{B7E1CD}1.24          & \cellcolor[HTML]{B7E1CD}0.18          & \cellcolor[HTML]{B7E1CD}0.25          & \cellcolor[HTML]{B7E1CD}\textbf{6.17} & \cellcolor[HTML]{B7E1CD}3.86          & \cellcolor[HTML]{B7E1CD}3.74          & \cellcolor[HTML]{B7E1CD}2.4          & \cellcolor[HTML]{B7E1CD}0.31          & \cellcolor[HTML]{B7E1CD}0.23          & \cellcolor[HTML]{F4C7C3}-0.35         & \cellcolor[HTML]{B7E1CD}0.33          & \cellcolor[HTML]{B7E1CD}0.17         & \cellcolor[HTML]{F4C7C3}-0.27         & \cellcolor[HTML]{F4C7C3}-0.41         & \cellcolor[HTML]{F4C7C3}-0.95         & \cellcolor[HTML]{F4C7C3}-0.47         & \cellcolor[HTML]{F4C7C3}-0.57         & \cellcolor[HTML]{B7E1CD}1.41          \\ \hline
\textbf{B+Mosaic}                 & \cellcolor[HTML]{B7E1CD}2.77          & \cellcolor[HTML]{B7E1CD}3.82          & \cellcolor[HTML]{B7E1CD}3.97          & \cellcolor[HTML]{B7E1CD}1.45          & \cellcolor[HTML]{B7E1CD}2.74          & \cellcolor[HTML]{B7E1CD}2.29          & \cellcolor[HTML]{B7E1CD}2.79         & \cellcolor[HTML]{B7E1CD}0.54          & \cellcolor[HTML]{B7E1CD}0.48          & \cellcolor[HTML]{B7E1CD}0.18          & \cellcolor[HTML]{B7E1CD}0.36          & \cellcolor[HTML]{B7E1CD}0.44         & \cellcolor[HTML]{F4C7C3}-0.21         & \cellcolor[HTML]{F4C7C3}-0.5          & \cellcolor[HTML]{B7E1CD}1.2           & \cellcolor[HTML]{B7E1CD}1.56          & \cellcolor[HTML]{F4C7C3}-0.27         & \cellcolor[HTML]{B7E1CD}1.28          \\ \hline
\textbf{B+pixAug}                 & \cellcolor[HTML]{B7E1CD}0.78          & \cellcolor[HTML]{B7E1CD}0.01          & \cellcolor[HTML]{B7E1CD}0.05          & \cellcolor[HTML]{B7E1CD}5.53          & \cellcolor[HTML]{B7E1CD}4.12          & \cellcolor[HTML]{B7E1CD}\textbf{3.97} & \cellcolor[HTML]{B7E1CD}3.4          & \cellcolor[HTML]{B7E1CD}0.56          & \cellcolor[HTML]{B7E1CD}0.44          & \cellcolor[HTML]{B7E1CD}0.88          & \cellcolor[HTML]{B7E1CD}0.82          & \cellcolor[HTML]{B7E1CD}0.42         & \cellcolor[HTML]{F4C7C3}-3.55         & \cellcolor[HTML]{F4C7C3}-3.31         & \cellcolor[HTML]{F4C7C3}-2.02         & \cellcolor[HTML]{F4C7C3}-2.73         & \cellcolor[HTML]{F4C7C3}-3.49         & \cellcolor[HTML]{F4C7C3}-1.99         \\ \hline
\textbf{B+pixAug+mosaic}          & \cellcolor[HTML]{F4C7C3}-1.92         & \cellcolor[HTML]{F4C7C3}-1.85         & \cellcolor[HTML]{F4C7C3}-0.64         & \cellcolor[HTML]{B7E1CD}1.89          & \cellcolor[HTML]{B7E1CD}2.79          & \cellcolor[HTML]{B7E1CD}2.25          & \cellcolor[HTML]{B7E1CD}\textbf{4.1} & \cellcolor[HTML]{B7E1CD}\textbf{1.52} & \cellcolor[HTML]{B7E1CD}0.95          & \cellcolor[HTML]{B7E1CD}1.01          & \cellcolor[HTML]{B7E1CD}\textbf{1.18} & \cellcolor[HTML]{B7E1CD}0.66         & \cellcolor[HTML]{F4C7C3}-7.27         & \cellcolor[HTML]{F4C7C3}-5.67         & \cellcolor[HTML]{F4C7C3}-3.72         & \cellcolor[HTML]{F4C7C3}-6.92         & \cellcolor[HTML]{F4C7C3}-5.78         & \cellcolor[HTML]{F4C7C3}-3.4          \\ \hline
\textbf{B+Box-Mixup (w/o iou)}    & \cellcolor[HTML]{B7E1CD}\textbf{4.59} & \cellcolor[HTML]{B7E1CD}\textbf{4.42} & \cellcolor[HTML]{B7E1CD}\textbf{4.42} & \cellcolor[HTML]{B7E1CD}4.99          & \cellcolor[HTML]{B7E1CD}\textbf{5.41} & \cellcolor[HTML]{B7E1CD}3.83          & \cellcolor[HTML]{B7E1CD}1.76         & \cellcolor[HTML]{F4C7C3}-0.98         & \cellcolor[HTML]{F4C7C3}-1.6          & \cellcolor[HTML]{F4C7C3}-0.89         & \cellcolor[HTML]{F4C7C3}-1.2          & \cellcolor[HTML]{F4C7C3}-1.46        & \cellcolor[HTML]{B7E1CD}\textbf{4.29} & \cellcolor[HTML]{B7E1CD}\textbf{1.39} & \cellcolor[HTML]{B7E1CD}\textbf{2.32} & \cellcolor[HTML]{B7E1CD}\textbf{4.94} & \cellcolor[HTML]{B7E1CD}\textbf{1.89} & \cellcolor[HTML]{B7E1CD}\textbf{2.83} \\ \hline
\textbf{B+Box-Mixup (w iou)}      & \cellcolor[HTML]{B7E1CD}3.42          & \cellcolor[HTML]{B7E1CD}3.9           & \cellcolor[HTML]{B7E1CD}4.2           & \cellcolor[HTML]{B7E1CD}3.15          & \cellcolor[HTML]{B7E1CD}3             & \cellcolor[HTML]{B7E1CD}3.03          & \cellcolor[HTML]{B7E1CD}3.64         & \cellcolor[HTML]{B7E1CD}1.16          & \cellcolor[HTML]{B7E1CD}\textbf{1.01} & \cellcolor[HTML]{B7E1CD}\textbf{1.18} & \cellcolor[HTML]{B7E1CD}1.07          & \cellcolor[HTML]{B7E1CD}\textbf{1.3} & \cellcolor[HTML]{F4C7C3}-1.31         & \cellcolor[HTML]{F4C7C3}-1.98         & \cellcolor[HTML]{B7E1CD}0.04          & \cellcolor[HTML]{F4C7C3}-1.44         & \cellcolor[HTML]{F4C7C3}-1.89         & \cellcolor[HTML]{B7E1CD}0.22          \\ \hline
\textbf{B+Box-CutPaste (w/o iou)} & \cellcolor[HTML]{B7E1CD}2.35          & \cellcolor[HTML]{B7E1CD}3.87          & \cellcolor[HTML]{B7E1CD}4.15          & \cellcolor[HTML]{B7E1CD}1.49          & \cellcolor[HTML]{B7E1CD}2.83          & \cellcolor[HTML]{B7E1CD}2.98          & \cellcolor[HTML]{B7E1CD}3.51         & \cellcolor[HTML]{B7E1CD}0.59          & \cellcolor[HTML]{B7E1CD}0.58          & \cellcolor[HTML]{B7E1CD}0.98          & \cellcolor[HTML]{B7E1CD}0.51          & \cellcolor[HTML]{B7E1CD}0.83         & \cellcolor[HTML]{B7E1CD}2.51          & \cellcolor[HTML]{B7E1CD}0.55          & \cellcolor[HTML]{B7E1CD}1.42          & \cellcolor[HTML]{B7E1CD}2.01          & \cellcolor[HTML]{B7E1CD}0.36          & \cellcolor[HTML]{B7E1CD}1.91          \\ \hline
\textbf{B+Box-CutPaste (w/ iou)}  & \cellcolor[HTML]{B7E1CD}4.08          & \cellcolor[HTML]{B7E1CD}4.17          & \cellcolor[HTML]{B7E1CD}4.27          & \cellcolor[HTML]{B7E1CD}3.63          & \cellcolor[HTML]{B7E1CD}3.3           & \cellcolor[HTML]{B7E1CD}3.42          & \cellcolor[HTML]{B7E1CD}3.58         & \cellcolor[HTML]{B7E1CD}1.21          & \cellcolor[HTML]{B7E1CD}0.72          & \cellcolor[HTML]{B7E1CD}1             & \cellcolor[HTML]{B7E1CD}1             & \cellcolor[HTML]{B7E1CD}0.72         & \cellcolor[HTML]{B7E1CD}3.54          & \cellcolor[HTML]{B7E1CD}0.94          & \cellcolor[HTML]{B7E1CD}1.92          & \cellcolor[HTML]{B7E1CD}3.03          & \cellcolor[HTML]{B7E1CD}0.69          & \cellcolor[HTML]{B7E1CD}1.93          \\ \hline
\end{tabular}
\end{table*}

We evaluate our data augmentations on the KITTI 3D detection benchmark which consists of 7,481 labeled training samples and 7518 unlabeled testing samples. Since the ground truth labels for the test set are not available, we evaluated our model by splitting the training set into 3711 training samples and 3768 validation samples. We experiment with ResNet-18 as backbone. We implemented our deep neural network in Pytorch and trained using Adam optimizer with learning rate of 1.25*1e-4 for 200 epochs. We trained our network with a batch size of 16. Our model achieved best speed with 33 FPS on a NVIDIA GTX 2080Ti GPU. 
\subsection{Results}
The KITTI benchmark evaluates the models by Average Precision(AP) of each class (Car, Pedestrain and Cyclist) under easy, moderate and hard conditions. We report two official evaluation metrics for “Test set” for comprehensive evaluation: AP for 3D bounding boxes 
$AP_{3d}$ and  AP for Birds Eye View $AP_{bev}$. We tabulate the results under two IoU thresholds for each class. IoU = 0.7 and IoU = 0.5 for Car category. IoU = 0.5 and IoU = 0.25 for Pedestrian and Cyclist categories. Comparison of baseline results with our proposed data augmentations is shown in Table \ref{tab:results}. IoU thresholds for Car are 0.7 while  0.25 for pedestrian and cyclists.
\begin{table*}[htb!]
\centering
\caption{mAP and ICFW mAP scores for both 3D and BEV detection bounding boxes at various IoUs. The values of different data augmentations below the baseline correspond to the difference. Green refers to positive gains, while red refers to negative drops in performance.}
\label{tab:mAP}
\begin{tabular}{|l|c|c|c|c|c|c|c|c|c|c|c|c|}
\hline
\multicolumn{1}{|c|}{\cellcolor[HTML]{FFFFFF}\textbf{IoU = 0.25}} & \multicolumn{3}{c|}{mAP$_{3D}$}                                                                                                                          & \multicolumn{3}{c|}{mAP$_{BEV}$}                                                                                                                        & \multicolumn{3}{c|}{ICFW mAP$_{3D}$}                                                                                                                       & \multicolumn{3}{c|}{ICFW mAP$_{BEV}$}                                                                                                                   \\ \hline
                                                              & E                                               & M                                               & H                                               & E                                               & M                                               & H                                               & E                                               & M                                               & H                                               & E                                               & M                                               & H                                               \\ \hline
Baseline (B)                                                 & \multicolumn{1}{r|}{\cellcolor[HTML]{CCCCCC}32.60} & \multicolumn{1}{r|}{\cellcolor[HTML]{CCCCCC}25.63} & \multicolumn{1}{r|}{\cellcolor[HTML]{CCCCCC}24.57} & \multicolumn{1}{r|}{\cellcolor[HTML]{CCCCCC}40.87} & \multicolumn{1}{r|}{\cellcolor[HTML]{CCCCCC}32.74} & \multicolumn{1}{r|}{\cellcolor[HTML]{CCCCCC}29.19} & \multicolumn{1}{r|}{\cellcolor[HTML]{CCCCCC}15.16} & \multicolumn{1}{r|}{\cellcolor[HTML]{CCCCCC}19.63} & \multicolumn{1}{r|}{\cellcolor[HTML]{CCCCCC}16.59} & \multicolumn{1}{r|}{\cellcolor[HTML]{CCCCCC}27.03} & \multicolumn{1}{r|}{\cellcolor[HTML]{CCCCCC}19.63} & \multicolumn{1}{r|}{\cellcolor[HTML]{CCCCCC}16.59} \\ \hline
B+Box-Mixup (w/o iou)                                          & \cellcolor[HTML]{F4C7C3}-0.50                   & \cellcolor[HTML]{B7E1CD}0.80                    & \cellcolor[HTML]{F4C7C3}-0.83                   & \cellcolor[HTML]{B7E1CD}1.43                    & \cellcolor[HTML]{B7E1CD}0.56                    & \cellcolor[HTML]{B7E1CD}0.78                    & \cellcolor[HTML]{B7E1CD}1.05                    & \cellcolor[HTML]{B7E1CD}1.80                    & \cellcolor[HTML]{B7E1CD}1.85                    & \cellcolor[HTML]{B7E1CD}3.51                    & \cellcolor[HTML]{B7E1CD}1.80                    & \cellcolor[HTML]{B7E1CD}1.85                    \\ \hline
B+Box-Mixup (w iou)                                            & \cellcolor[HTML]{F4C7C3}-0.10                   & \cellcolor[HTML]{B7E1CD}1.89                    & \cellcolor[HTML]{B7E1CD}0.29                    & \cellcolor[HTML]{F4C7C3}-0.26                   & \cellcolor[HTML]{F4C7C3}-0.32                   & \cellcolor[HTML]{B7E1CD}0.50                    & \cellcolor[HTML]{B7E1CD}0.09                    & \cellcolor[HTML]{F4C7C3}-1.74                   & \cellcolor[HTML]{B7E1CD}0.44                    & \cellcolor[HTML]{F4C7C3}-0.88                   & \cellcolor[HTML]{F4C7C3}-1.74                   & \cellcolor[HTML]{B7E1CD}0.44                    \\ \hline
B+Box-CutPaste (w/o iou)                                       & \cellcolor[HTML]{F4C7C3}-6.17                   & \cellcolor[HTML]{F4C7C3}-3.60                   & \cellcolor[HTML]{F4C7C3}-5.32                   & \cellcolor[HTML]{F4C7C3}-8.19                   & \cellcolor[HTML]{F4C7C3}-5.50                   & \cellcolor[HTML]{B7E1CD}0.84                    & \cellcolor[HTML]{F4C7C3}-2.84                   & \cellcolor[HTML]{B7E1CD}0.31                    & \cellcolor[HTML]{B7E1CD}1.60                    & \cellcolor[HTML]{B7E1CD}1.53                    & \cellcolor[HTML]{B7E1CD}0.31                    & \cellcolor[HTML]{B7E1CD}1.60                    \\ \hline
B+Box-CutPaste (w/ iou)                                        & \cellcolor[HTML]{F4C7C3}-0.50                   & \cellcolor[HTML]{B7E1CD}1.84                    & \cellcolor[HTML]{B7E1CD}0.22                    & \cellcolor[HTML]{B7E1CD}0.88                    & \cellcolor[HTML]{B7E1CD}0.54                    & \cellcolor[HTML]{B7E1CD}0.91                    & \cellcolor[HTML]{B7E1CD}0.15                    & \cellcolor[HTML]{B7E1CD}0.65                    & \cellcolor[HTML]{B7E1CD}1.60                    & \cellcolor[HTML]{B7E1CD}2.33                    & \cellcolor[HTML]{B7E1CD}0.65                    & \cellcolor[HTML]{B7E1CD}1.60                    \\ \hline
\multicolumn{1}{|c|}{\cellcolor[HTML]{FFFFFF}\textbf{IoU = 0.5}}  & \multicolumn{3}{c|}{mAP$_{3D}$}                                                                                                                          & \multicolumn{3}{c|}{mAP$_{BEV}$}                                                                                                        & \multicolumn{3}{c|}{ICFW mAP$_{3D}$}                                                                                                                     & \multicolumn{3}{c|}{ICFW mAP$_{BEV}$}                                                                                                                   \\ \hline
                                                              & E                                               & M                                               & H                                               & E                                               & M                                               & H                                               & E                                               & M                                               & H                                               & E                                               & M                                               & H                                               \\ \hline
Baseline (B)                                                   & \cellcolor[HTML]{CCCCCC}22.78                      & \cellcolor[HTML]{CCCCCC}17.80                      & \cellcolor[HTML]{CCCCCC}16.85                      & \cellcolor[HTML]{CCCCCC}33.39                      & \cellcolor[HTML]{CCCCCC}25.08                      & \cellcolor[HTML]{CCCCCC}22.23                      & \cellcolor[HTML]{CCCCCC}13.05                      & \cellcolor[HTML]{CCCCCC}18.16                      & \cellcolor[HTML]{CCCCCC}15.73                      & \cellcolor[HTML]{CCCCCC}25.42                      & \cellcolor[HTML]{CCCCCC}18.16                      & \cellcolor[HTML]{CCCCCC}15.73                      \\ \hline
B+Box-Mixup (w/o iou)                                          & \cellcolor[HTML]{B7E1CD}0.04                    & \cellcolor[HTML]{F4C7C3}-0.59                   & \cellcolor[HTML]{F4C7C3}-0.78                   & \cellcolor[HTML]{B7E1CD}1.80                    & \cellcolor[HTML]{B7E1CD}0.62                    & \cellcolor[HTML]{B7E1CD}0.71                    & \cellcolor[HTML]{B7E1CD}1.17                    & \cellcolor[HTML]{B7E1CD}1.81                    & \cellcolor[HTML]{B7E1CD}1.85                    & \cellcolor[HTML]{B7E1CD}3.59                    & \cellcolor[HTML]{B7E1CD}1.81                    & \cellcolor[HTML]{B7E1CD}1.85                    \\ \hline
B+Box-Mixup (w iou)                                            & \cellcolor[HTML]{B7E1CD}0.04                    & \cellcolor[HTML]{B7E1CD}0.41                    & \cellcolor[HTML]{B7E1CD}0.21                    & \cellcolor[HTML]{F4C7C3}-0.11                   & \cellcolor[HTML]{F4C7C3}-0.31                   & \cellcolor[HTML]{B7E1CD}0.44                    & \cellcolor[HTML]{B7E1CD}0.12                    & \cellcolor[HTML]{F4C7C3}-1.74                   & \cellcolor[HTML]{B7E1CD}0.43                    & \cellcolor[HTML]{F4C7C3}-0.85                   & \cellcolor[HTML]{F4C7C3}-1.74                   & \cellcolor[HTML]{B7E1CD}0.43                    \\ \hline
B+Box-CutPaste (w/o iou)                                       & \cellcolor[HTML]{F4C7C3}-6.52                   & \cellcolor[HTML]{F4C7C3}-5.29                   & \cellcolor[HTML]{F4C7C3}-5.50                   & \cellcolor[HTML]{F4C7C3}-9.26                   & \cellcolor[HTML]{F4C7C3}-5.75                   & \cellcolor[HTML]{B7E1CD}0.69                    & \cellcolor[HTML]{F4C7C3}-2.91                   & \cellcolor[HTML]{B7E1CD}0.26                    & \cellcolor[HTML]{B7E1CD}1.58                    & \cellcolor[HTML]{B7E1CD}1.30                    & \cellcolor[HTML]{B7E1CD}0.26                    & \cellcolor[HTML]{B7E1CD}1.58                    \\ \hline
B+Box-CutPaste (w/ iou)                                        & \cellcolor[HTML]{F4C7C3}-0.39                   & \cellcolor[HTML]{B7E1CD}0.39                    & \cellcolor[HTML]{B7E1CD}0.23                    & \cellcolor[HTML]{B7E1CD}0.45                    & \cellcolor[HTML]{B7E1CD}0.42                    & \cellcolor[HTML]{B7E1CD}0.71                    & \cellcolor[HTML]{B7E1CD}0.17                    & \cellcolor[HTML]{B7E1CD}0.62                    & \cellcolor[HTML]{B7E1CD}1.58                    & \cellcolor[HTML]{B7E1CD}2.24                    & \cellcolor[HTML]{B7E1CD}0.62                    & \cellcolor[HTML]{B7E1CD}1.58                    \\ \hline
\multicolumn{1}{|c|}{\cellcolor[HTML]{FFFFFF}\textbf{IoU = 0.7}}  & \multicolumn{3}{c|}{mAP$_{3D}$}                                                                                                                          & \multicolumn{3}{c|}{mAP$_{BEV}$}                                                                                                                        & \multicolumn{3}{c|}{ICFW mAP$_{3D}$}                                                                                                                       & \multicolumn{3}{c|}{ICFW mAP$_{BEV}$}                                                                                                                   \\ \hline
                                                              & E                                               & M                                               & H                                               & E                                               & M                                               & H                                               & E                                               & M                                               & H                                               & E                                               & M                                               & H                                               \\ \hline
Baseline (B)                                                   & \multicolumn{1}{r|}{\cellcolor[HTML]{CCCCCC}4.53}  & \multicolumn{1}{r|}{\cellcolor[HTML]{CCCCCC}3.67}  & \multicolumn{1}{r|}{\cellcolor[HTML]{CCCCCC}3.07}  & \multicolumn{1}{r|}{\cellcolor[HTML]{CCCCCC}6.73}  & \multicolumn{1}{r|}{\cellcolor[HTML]{CCCCCC}5.20}  & \multicolumn{1}{r|}{\cellcolor[HTML]{CCCCCC}4.93}  & \multicolumn{1}{r|}{\cellcolor[HTML]{CCCCCC}1.54}  & \multicolumn{1}{r|}{\cellcolor[HTML]{CCCCCC}1.36}  & \multicolumn{1}{r|}{\cellcolor[HTML]{CCCCCC}1.08}  & \multicolumn{1}{r|}{\cellcolor[HTML]{CCCCCC}2.12}  & \multicolumn{1}{r|}{\cellcolor[HTML]{CCCCCC}1.79}  & \multicolumn{1}{r|}{\cellcolor[HTML]{CCCCCC}1.74}  \\ \hline
B+Box-Mixup (w/o iou)                                          & \cellcolor[HTML]{B7E1CD}1.16                    & \cellcolor[HTML]{B7E1CD}1.10                    & \cellcolor[HTML]{B7E1CD}1.18                    & \cellcolor[HTML]{B7E1CD}1.37                    & \cellcolor[HTML]{B7E1CD}1.48                    & \cellcolor[HTML]{B7E1CD}0.96                    & \cellcolor[HTML]{F4C7C3}-0.26                   & \cellcolor[HTML]{F4C7C3}-0.27                   & \cellcolor[HTML]{F4C7C3}-0.10                   & \cellcolor[HTML]{F4C7C3}-0.04                   & \cellcolor[HTML]{F4C7C3}-0.09                   & \cellcolor[HTML]{F4C7C3}-0.20                   \\ \hline
B+Box-Mixup (w iou)                                            & \cellcolor[HTML]{B7E1CD}1.49                    & \cellcolor[HTML]{B7E1CD}1.52                    & \cellcolor[HTML]{B7E1CD}1.63                    & \cellcolor[HTML]{B7E1CD}1.36                    & \cellcolor[HTML]{B7E1CD}1.25                    & \cellcolor[HTML]{B7E1CD}1.29                    & \cellcolor[HTML]{B7E1CD}0.51                    & \cellcolor[HTML]{B7E1CD}0.52                    & \cellcolor[HTML]{B7E1CD}0.69                    & \cellcolor[HTML]{B7E1CD}0.40                    & \cellcolor[HTML]{B7E1CD}0.37                    & \cellcolor[HTML]{B7E1CD}0.38                    \\ \hline
B+Box-CutPaste (w/o iou)                                       & \cellcolor[HTML]{B7E1CD}0.92                    & \cellcolor[HTML]{B7E1CD}1.06                    & \cellcolor[HTML]{B7E1CD}1.23                    & \cellcolor[HTML]{B7E1CD}1.14                    & \cellcolor[HTML]{B7E1CD}1.01                    & \cellcolor[HTML]{B7E1CD}1.02                    & \cellcolor[HTML]{F4C7C3}-0.08                   & \cellcolor[HTML]{F4C7C3}-0.06                   & \cellcolor[HTML]{B7E1CD}0.12                    & \cellcolor[HTML]{B7E1CD}0.61                    & \cellcolor[HTML]{B7E1CD}0.44                    & \cellcolor[HTML]{B7E1CD}0.44                    \\ \hline
B+Box-CutPaste (w/ iou)                                        & \cellcolor[HTML]{B7E1CD}0.76                    & \cellcolor[HTML]{B7E1CD}1.12                    & \cellcolor[HTML]{B7E1CD}1.29                    & \cellcolor[HTML]{B7E1CD}0.62                    & \cellcolor[HTML]{B7E1CD}0.87                    & \cellcolor[HTML]{B7E1CD}0.92                    & \cellcolor[HTML]{B7E1CD}0.15                    & \cellcolor[HTML]{F4C7C3}-0.07                   & \cellcolor[HTML]{B7E1CD}0.12                    & \cellcolor[HTML]{B7E1CD}0.41                    & \cellcolor[HTML]{B7E1CD}0.08                    & \cellcolor[HTML]{B7E1CD}0.09                    \\ \hline
\end{tabular}
\end{table*}

To further describe the results we use two other evaluation metrics already known in literature: First, the Mean Average Precision (mAP) The mean value of the Average Precision (AP) over all classes : 
\begin{equation}
    \text{mAP}_{3D} = \frac{1}{\left|C\right|} \sum_{c \in C} \text{AP}_c
\end{equation}
where $C=\{\text{car, pedestrian, cyclist}\}$.
Second we introduce the, Inverse Class Frequency Weighted (ICFW) mAP metric. This new metric was introduced to demonstrate the class imbalance inherent in the KITTI dataset. The relative frequency (denoted by $f_c$ and in blue) of car, pedestrian and cyclist classes in the validation classes for the different difficulty levels (based on thresholds over object occlusion levels) are shown in Table \ref{tab:classfreq}. To reduce the effect of the dominant class (here the car) and observe which data augmentation method produces gains in detection in least represented class, we introduce the inverse class weighting. This is evaluated by the following formula : 
\begin{equation}
    w_c := \frac{f_c^{-1}}{\sum_{c \in C} f_c^{-1}} \in [0, 1] \text{ \ and \ } 
    \sum_{c \in C} w_c = 1
\end{equation}
The values of $w_c$ are shown in Table \ref{tab:classfreq} in red.
Now the ICFW mAP is evaluated as :
\begin{equation}
    \text{ICFW mAP$_{3D}$} =  \sum_{c \in C} w_c \text{AP}_c
\end{equation}

\subsection{Analysis}

From Table \ref{tab:results}, it is clear that Box - Mixup outperforms all other augmentations for “Car” class in terms of $AP_{3d}$ and $AP_{bev}$ for both IoU = 0.7 and 0.5. Cutout-2 holes is best suited augmentation for “Pedestrian” class for $AP_{3d}(IoU=0.5)$ and $AP_{bev}(IoU=0.5  \&  0.25)$. For “Cyclist” class, different augmentations suit different conditions.\\
Table \ref{tab:mAP} shows different combinations of the proposed augmentations experimented on KITTI dataset. It tabulates the average mAP for each difficulty level (E, M and H) over all classes for IoU = 0.5. 
The table shows in red cases where the data augmentation performed poorly, and in green where it performed better, with each entry showing the difference w.r.t the baseline.

We observe that even though the mAP scores improvements might be negligible or negative in for some data augmentations the ICFW mAP scores reflect the improvement in performance reweighted due to reflect poor pedestrian and cyclist class frequencies. 
\begin{figure*}[hbt!]
\setlength{\lineskip}{0pt}
\centering
\begin{subfigure}[b]{0.45\textwidth}
\centering
    Baseline model\\
    \vspace{1mm}
    \includegraphics[width=\textwidth]{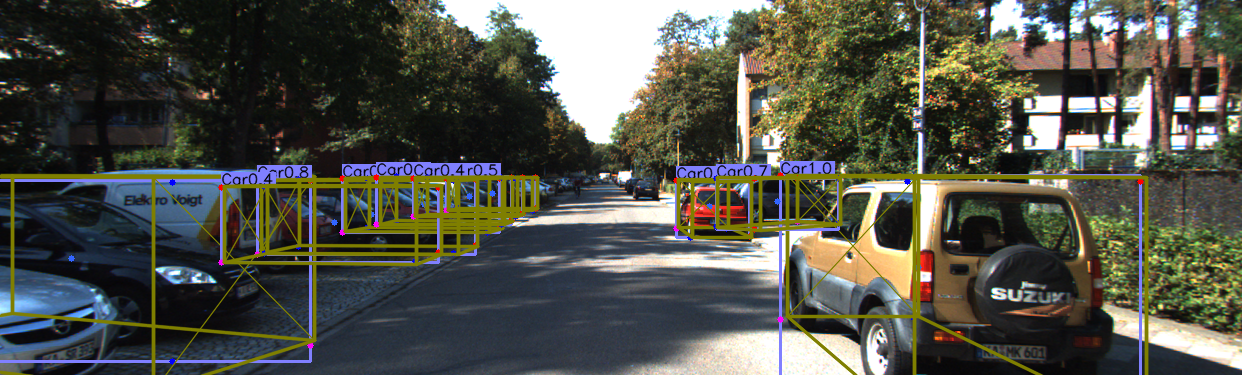}
    BoxMixup Augmentation\\
    \vspace{1mm}
    \includegraphics[width=\textwidth]{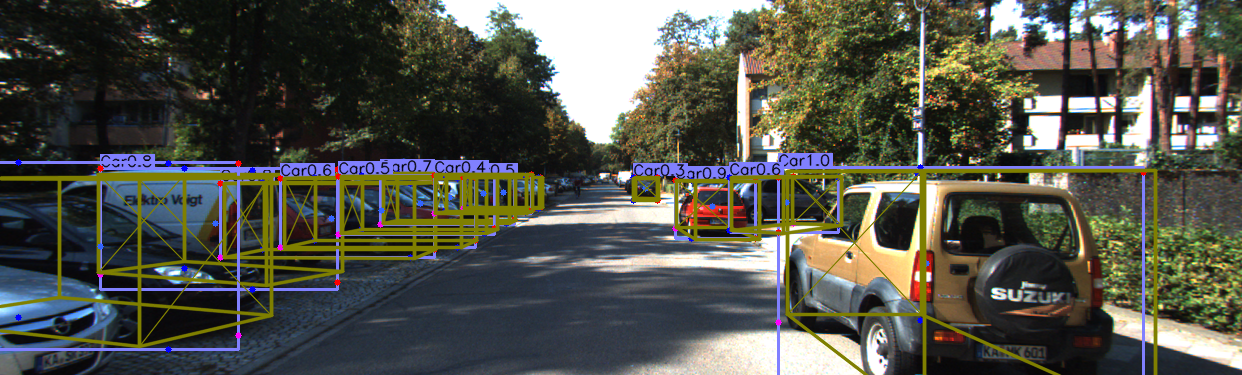}
\end{subfigure}
\hfill
\begin{subfigure}[b]{0.45\textwidth}
\centering
    \includegraphics[width=0.375\textwidth]{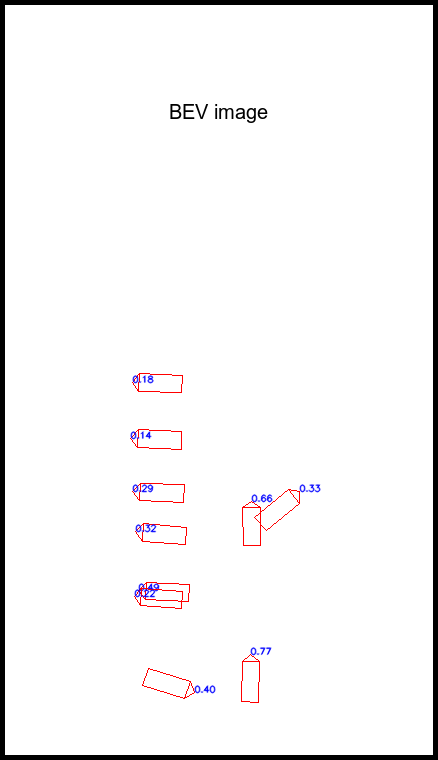}
    \includegraphics[width=0.375\textwidth]{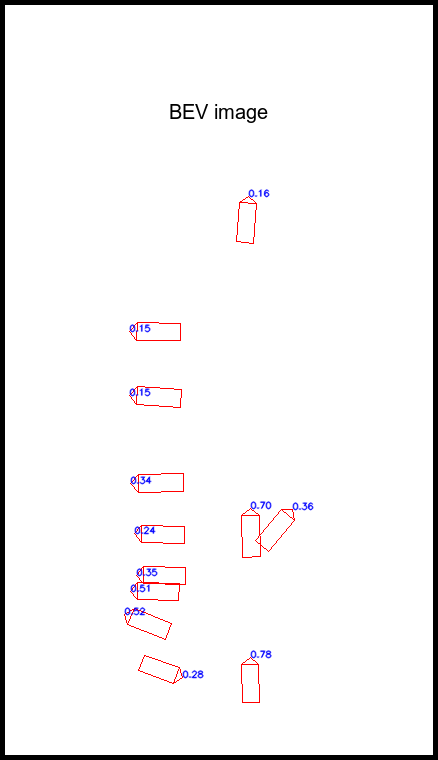}
    \caption{Left: Baseline, Right : Box-MixUp}
\end{subfigure}
\hfill
\begin{subfigure}[b]{0.45\textwidth}
\centering
    \vspace{1mm}
    Baseline model\\
    \vspace{1mm}
    \includegraphics[width=\textwidth]{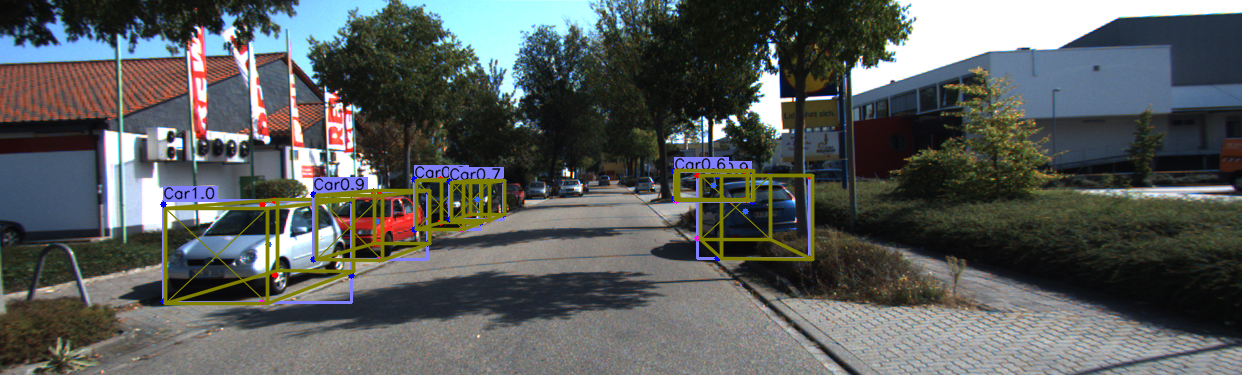}
    BoxMixup Augmentation\\
    \vspace{1mm}
    \includegraphics[width=\textwidth]{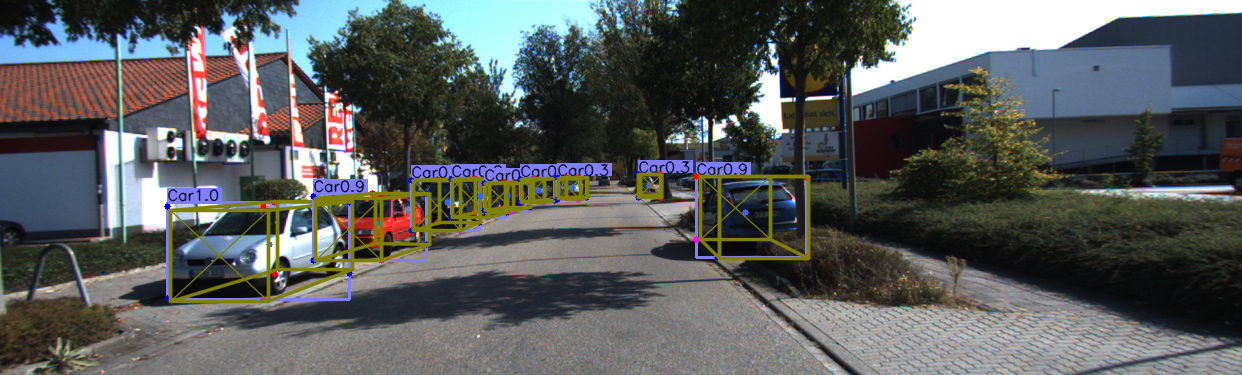}
\end{subfigure}
\hfill
\begin{subfigure}[b]{0.45\textwidth}
\centering
    \includegraphics[width=0.375\textwidth]{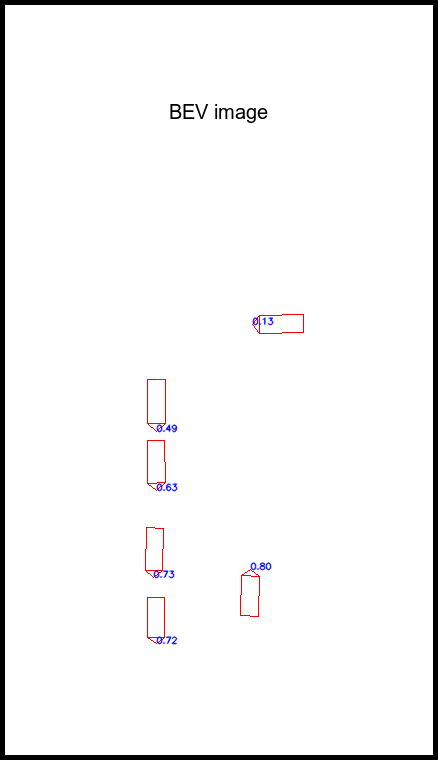}
    \includegraphics[width=0.375\textwidth]{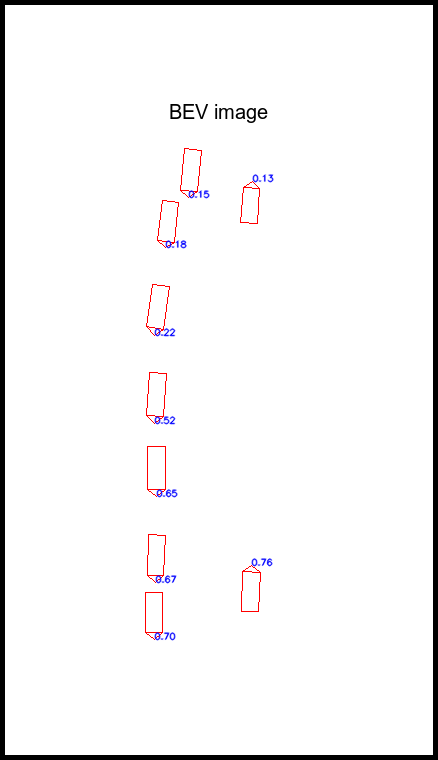}
    \caption{Left: Baseline, Right : Box-MixUp}
\end{subfigure}
\hfill
\begin{subfigure}[b]{0.45\textwidth}
\centering
    \vspace{1mm}    
    Baseline model\\
    \vspace{1mm}
    \includegraphics[width=\textwidth]{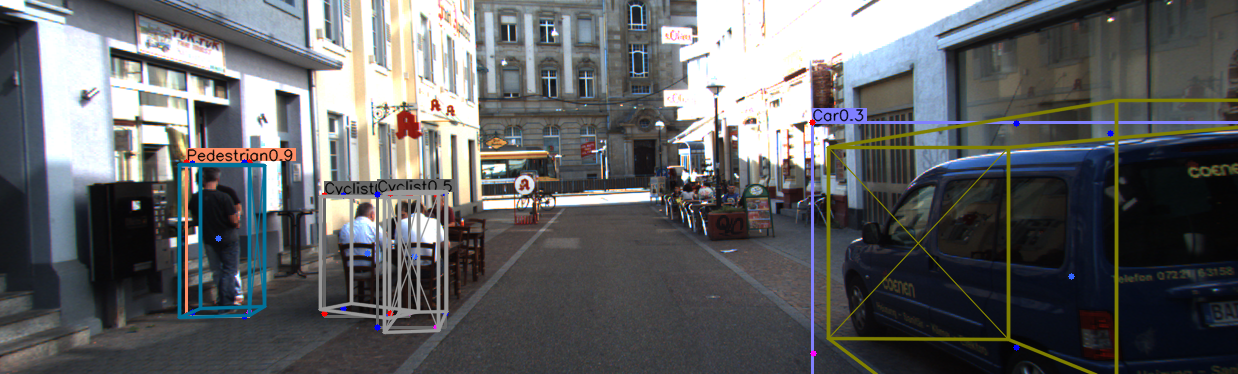}
    BoxMixup Augmentation\\
    \vspace{1mm}
    \includegraphics[width=\textwidth]{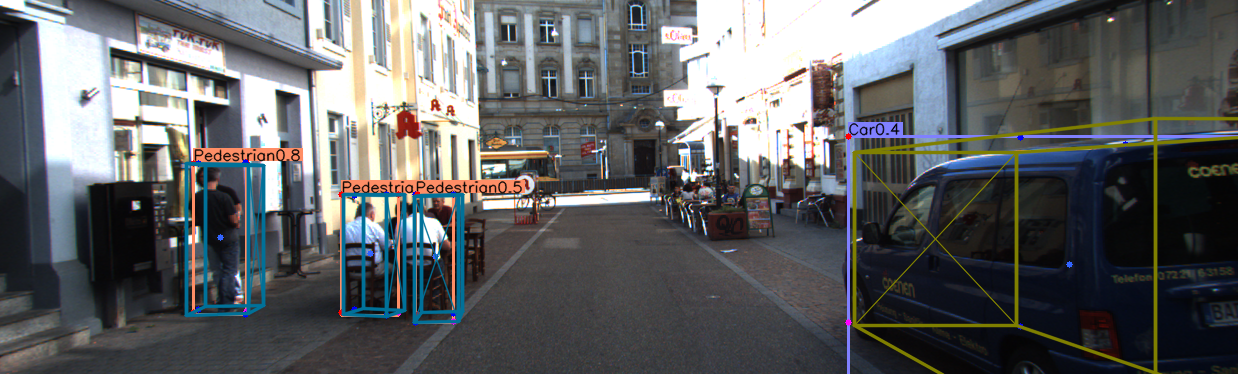}
\end{subfigure}
\hfill
\begin{subfigure}[b]{0.45\textwidth}
\centering
    \includegraphics[width=0.375\textwidth]{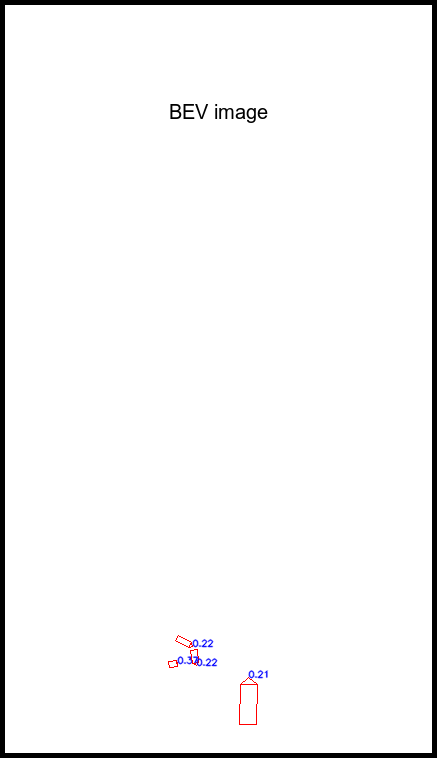}
    \includegraphics[width=0.375\textwidth]{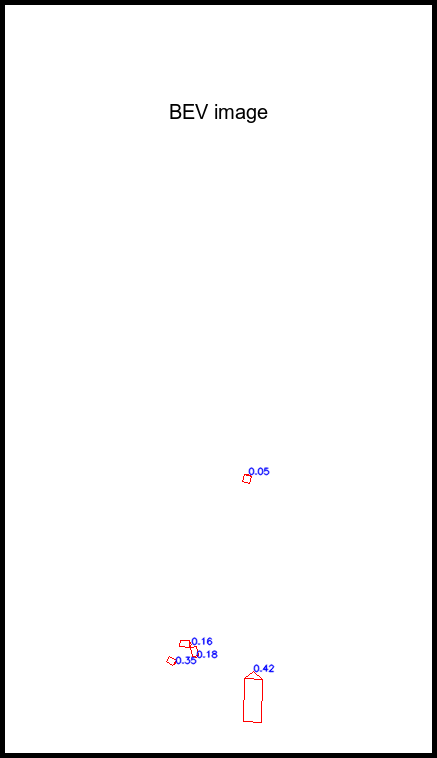}
    \caption{Left: Baseline, Right : Box-MixUp}
\end{subfigure}

\caption{Illustration of Box-Mixup data augmentation in various scenarios. Each time contains
the (baseline, Box-Mixup) prediction pair on the left panel, while the BEV representations (baseline, data augmented) pair on the right panel.}
\label{fig:img-pairs}
\end{figure*}

The mAP of the RTM3D detector is improved reasonably with our proposed data augmentations. We find that our proposed augmentations are applicable over any baseline independent of the method. The RTM3D has two main data augmentation schemes already implemented which we consider as our baseline model : 1. Affine transformations, 2. image transforms. We do not include the stereo image pair from the KITTI dataset to extend the size of the dataset, or cited by \cite{rtm3deccv2020} as stereo dataset augmentation. This produces a different baseline result in our case.

Fig. \ref{fig:img-pairs} shows the detection results of our proposed Box-Mixup data augmentation on KITTI in different scenarios. Ex. Occluded objects, missed detections and mis-classification by baseline. It shows baseline, Box-Mixup predictions on left panel and their corresponding BEV representations on right panel.\\

\subsection{Analysis of Results}
From the comparison in Table \ref{tab:mAP}, the key data augmentation 
where observed improvement were the Box-Mixup and Box-CutPaste with an 
IoU check. When used without the IoU check we observe the drop in ICFW 
mAP scores denoting that the car AP score dominated the mAP scores. 
Ensuring that no two boxes from different images have a large
intersection enables the data augmentation to provide a more diverse 
augmented images.

\begin{table}[bt!]
\caption{Class Frequencies over different difficulty cases Easy, Moderate, Hard with $(f_c/w_c)$ pair in each cell. }
\label{tab:classfreq}
\centering
\begin{tabular}{|l|c|c|c|}
\hline
\textbf{Class/Difficulty} & \textbf{Car} & \textbf{Pedestrian} & \textbf{Cyclist} \\
\hline
\textbf{Easy}     & \textcolor{blue}{0.68}/\textcolor{red}{0.07} & \textcolor{blue}{0.26}/\textcolor{red}{0.19} & \textcolor{blue}{0.07}/\textcolor{red}{0.74} \\
\textbf{Moderate} & \textcolor{blue}{0.78}/\textcolor{red}{0.06} & \textcolor{blue}{0.17}/\textcolor{red}{0.01} & \textcolor{blue}{0.05}/\textcolor{red}{0.92}  \\
\textbf{Hard}     & \textcolor{blue}{0.80}/\textcolor{red}{0.04} & \textcolor{blue}{0.16}/\textcolor{red}{0.21} & \textcolor{blue}{0.04}/\textcolor{red}{0.75} \\                     
\hline
\end{tabular}
\end{table}

\textbf{Effect of IoU : } Box-MixUp augmentations generated with low IoU constraint between training images, provide better performance for pedestrian and cyclists classes.

\textbf{Effect of Class imbalance : } The data augmentations that performed well on the majority class are not optimal for the low frequency pedestrian and cyclist classes.

Table \ref{tab:results} shows the class-wise performance of all data 
augmentation methods. Global evaluation of all proposed data 
augmentations is shown in Table \ref{tab:mAP}. Almost all data 
augmentations have increased performance over "Car" and "Cyclists" 
classes. But, performance gain across data augmentation methods on "Pedestrian" class is lower. We hypothesize that this is due to the symmetry in width and height of pedestrians (which is not the case with cyclists where there are 3 distinct dimensions). This makes the network uncertain in localizing "Pedestrian" class. 

\section{Conclusion}
We have evaluated a family of 2D data augmentations for monocular 3D object detection in images, along with a set of novel 2D data augmentations without changing the 3D geometry and avoids synthesizing new viewpoints. The data augmentations were applied over the RTM3D detector as baseline. In spite of these 2D data pixel/region level augmentations, improvements were observed in $(mAP_{3D})$ and $(mAP_{bev})$ scores. This was due to improved invariance to conditions like occlusion, corrupted pixels, novel orientations of object masks in 3D with varying background. As seen in Fig. \ref{fig:img-pairs} the Box-MixUp augmentation provides better separation of occluded vehicles, as well at a farther distance detection.

To achieve a better performance across all classes we aim to study a better sampling mechanism in future work that takes into account the class imbalance inherent in the KITTI dataset.

\bibliographystyle{IEEEtran}
\bibliography{IEEEabrv,bibliography}

\begin{thebibliography}{10}
\providecommand{\url}[1]{#1}
\csname url@rmstyle\endcsname
\providecommand{\newblock}{\relax}
\providecommand{\bibinfo}[2]{#2}
\providecommand\BIBentrySTDinterwordspacing{\spaceskip=0pt\relax}
\providecommand\BIBentryALTinterwordstretchfactor{4}
\providecommand\BIBentryALTinterwordspacing{\spaceskip=\fontdimen2\font plus
\BIBentryALTinterwordstretchfactor\fontdimen3\font minus
  \fontdimen4\font\relax}
\providecommand\BIBforeignlanguage[2]{{%
\expandafter\ifx\csname l@#1\endcsname\relax
\typeout{** WARNING: IEEEtran.bst: No hyphenation pattern has been}%
\typeout{** loaded for the language `#1'. Using the pattern for}%
\typeout{** the default language instead.}%
\else
\language=\csname l@#1\endcsname
\fi
#2}}

\bibitem{Law_2018_ECCV}
H.~Law and J.~Deng, ``Cornernet: Detecting objects as paired keypoints,'' in
  \emph{The European Conference on Computer Vision (ECCV)}, September 2018.

\bibitem{Zhou_2019_CVPR}
X.~Zhou, J.~Zhuo, and P.~Krahenbuhl, ``Bottom-up object detection by grouping
  extreme and center points,'' in \emph{The IEEE Conference on Computer Vision
  and Pattern Recognition (CVPR)}, June 2019.

\bibitem{Zhou2019ObjectsAP}
X.~Zhou, D.~Wang, and P.~Kr{\"a}henb{\"u}hl, ``Objects as points,''
  \emph{ArXiv}, vol. abs/1904.07850, 2019.

\bibitem{Li2020Monocular3D}
P.-X. Li, ``Monocular 3d detection with geometric constraints embedding and
  semi-supervised training,'' \emph{ArXiv}, vol. abs/2009.00764, 2020.

\bibitem{rtm3deccv2020}
P.~Li, H.~Zhao, P.~Liu, and F.~Cao, ``Rtm3d: Real-time monocular 3d detection
  from object keypoints for autonomous driving,'' in \emph{Computer Vision --
  ECCV 2020}, A.~Vedaldi, H.~Bischof, T.~Brox, and J.-M. Frahm, Eds.\hskip 1em
  plus 0.5em minus 0.4em\relax Cham: Springer International Publishing, 2020,
  pp. 644--660.

\bibitem{Redmon2018YOLOv3AI}
J.~Redmon and A.~Farhadi, ``Yolov3: An incremental improvement,'' \emph{ArXiv},
  vol. abs/1804.02767, 2018.

\bibitem{kittidataset2012}
A.~{Geiger}, P.~{Lenz}, and R.~{Urtasun}, ``Are we ready for autonomous
  driving? the kitti vision benchmark suite,'' in \emph{2012 IEEE Conference on
  Computer Vision and Pattern Recognition}, 2012, pp. 3354--3361.

\bibitem{liu2020smoke}
Z.~Liu, Z.~Wu, and R.~T{\'o}th, ``Smoke: single-stage monocular 3d object
  detection via keypoint estimation,'' in \emph{Proceedings of the IEEE/CVF
  Conference on Computer Vision and Pattern Recognition Workshops}, 2020, pp.
  996--997.

\bibitem{Chen_2016_CVPR}
X.~Chen, K.~Kundu, Z.~Zhang, H.~Ma, S.~Fidler, and R.~Urtasun, ``Monocular 3d
  object detection for autonomous driving,'' in \emph{Proceedings of the IEEE
  Conference on Computer Vision and Pattern Recognition (CVPR)}, June 2016.

\bibitem{Li_2019_CVPR}
B.~Li, W.~Ouyang, L.~Sheng, X.~Zeng, and X.~Wang, ``Gs3d: An efficient 3d
  object detection framework for autonomous driving,'' in \emph{The IEEE
  Conference on Computer Vision and Pattern Recognition (CVPR)}, June 2019.

\bibitem{Devries2017ImprovedRO}
T.~Devries and G.~W. Taylor, ``Improved regularization of convolutional neural
  networks with cutout,'' \emph{ArXiv}, vol. abs/1708.04552, 2017.

\bibitem{Zhang2018mixupBE}
H.~Zhang, M.~Cisse, Y.~N. Dauphin, and D.~Lopez-Paz, ``mixup: Beyond empirical
  risk minimization,'' \emph{International Conference on Learning
  Representations}, 2018.

\bibitem{yun2019cutmix}
S.~Yun, D.~Han, S.~J. Oh, S.~Chun, J.~Choe, and Y.~Yoo, ``Cutmix:
  Regularization strategy to train strong classifiers with localizable
  features,'' in \emph{Proceedings of the IEEE/CVF International Conference on
  Computer Vision}, 2019, pp. 6023--6032.

\bibitem{lee2020smoothmix}
J.-H. Lee, M.~Z. Zaheer, M.~Astrid, and S.-I. Lee, ``Smoothmix: a simple yet
  effective data augmentation to train robust classifiers,'' in
  \emph{Proceedings of the IEEE/CVF Conference on Computer Vision and Pattern
  Recognition Workshops}, 2020, pp. 756--757.

\bibitem{Bochkovskiy2020YOLOv4OS}
A.~Bochkovskiy, C.-Y. Wang, and H.~Liao, ``Yolov4: Optimal speed and accuracy
  of object detection,'' \emph{ArXiv}, vol. abs/2004.10934, 2020.

\bibitem{albumentations2020mdpi}
\BIBentryALTinterwordspacing
A.~Buslaev, V.~I. Iglovikov, E.~Khvedchenya, A.~Parinov, M.~Druzhinin, and
  A.~A. Kalinin, ``Albumentations: Fast and flexible image augmentations,''
  \emph{Information}, vol.~11, no.~2, 2020. [Online]. Available:
  \url{https://www.mdpi.com/2078-2489/11/2/125}
\BIBentrySTDinterwordspacing

\bibitem{manhardt2019roi}
F.~Manhardt, W.~Kehl, and A.~Gaidon, ``Roi-10d: Monocular lifting of 2d
  detection to 6d pose and metric shape,'' in \emph{Proceedings of the IEEE/CVF
  Conference on Computer Vision and Pattern Recognition}, 2019, pp. 2069--2078.

\bibitem{Zhang2020MultiModalityCA}
W.~Zhang, Z.~Wang, and C.~C. Loy, ``Multi-modality cut and paste for 3d object
  detection,'' \emph{ArXiv}, vol. abs/2012.12741, 2020.

\bibitem{Dwibedi2017CutPA}
D.~Dwibedi, I.~Misra, and M.~Hebert, ``Cut, paste and learn: Surprisingly easy
  synthesis for instance detection,'' \emph{2017 IEEE International Conference
  on Computer Vision (ICCV)}, pp. 1310--1319, 2017.

\bibitem{ravi2018real}
B.~Ravi~Kiran, L.~Roldao, B.~Irastorza, R.~Verastegui, S.~Suss, S.~Yogamani,
  V.~Talpaert, A.~Lepoutre, and G.~Trehard, ``Real-time dynamic object
  detection for autonomous driving using prior 3d-maps,'' in \emph{Proceedings
  of the European Conference on Computer Vision (ECCV) Workshops}, 2018.

\end{thebibliography}

\end{document}